 \documentclass[pmlr,twocolumn,10pt]{jmlr}

\usepackage{longtable}
\usepackage{csvsimple}
\usepackage{multirow}

 \usepackage{booktabs}

\usepackage[load-configurations=version-1]{siunitx} 

\usepackage{hyperref}
\usepackage{graphicx}
\usepackage{algorithm}

\jmlrproceedings{PMLR}{ML4H Findings Track Collection}
\jmlrworkshop{Machine Learning for Health (ML4H) 2023}

\usepackage{booktabs}
\usepackage{siunitx}

\title[XAIQA]{XAIQA}

 \title[XAIQA]{XAIQA: Explainer-Based Data Augmentation for Extractive Question Answering}

\author{\Name{Joel Stremmel}\Email{joel$\_$stremmel@optum.com}\\
  \Name{Ardavan Saeedi}\Email{ardavan.saeedi@optum.com}\\
  \Name{Hamid Hassanzadeh}\Email{hamid.hassanzadeh@optum.com}\\
  \Name{Sanjit Batra}\Email{sanjit.batra@optum.com}\\
  \Name{Jeffrey Hertzberg}\Email{jeffrey.hertzberg@optum.com}\\
  \Name{Jaime Murillo}\Email{jaime$\_$murillo@uhg.com}\\
  \Name{Eran Halperin}\Email{eran.halperin@optum.com}\\
  \addr Optum, Minnetonka, MN, USA}

\begin{document}

\maketitle

\begin{abstract}

 Extractive question answering (QA) systems can enable physicians and researchers to query medical records,
 a foundational capability for designing clinical studies and understanding patient medical history.
 However, building these systems typically requires expert-annotated QA pairs.
 Large language models (LLMs), which can perform extractive QA, depend on high quality data in their prompts,
 specialized for the application domain. We introduce a novel approach, XAIQA, for generating synthetic QA pairs
 at scale from data naturally available in electronic health records. 
 Our method uses the idea of a classification model explainer to generate questions and answers about medical concepts corresponding to medical codes.
 In an expert evaluation with two physicians, our method identifies $2.2\times$ more semantic matches and $3.8\times$ more clinical
 abbreviations than two popular approaches that use sentence transformers to create QA pairs.
 In an ML evaluation, adding our QA pairs improves performance of GPT-4 as an extractive QA model,
 including on difficult questions.
 In both the expert and ML evaluations, we examine trade-offs between our method and 
 sentence transformers for QA pair generation depending on question difficulty.

\end{abstract}

\begin{keywords}
Language Models, Transformers, Question Answering, Synthetic Data, Explainability
\end{keywords}

\section{Introduction}


\begin{figure*}[h]
  \centering
  \includegraphics[width=4.5in]{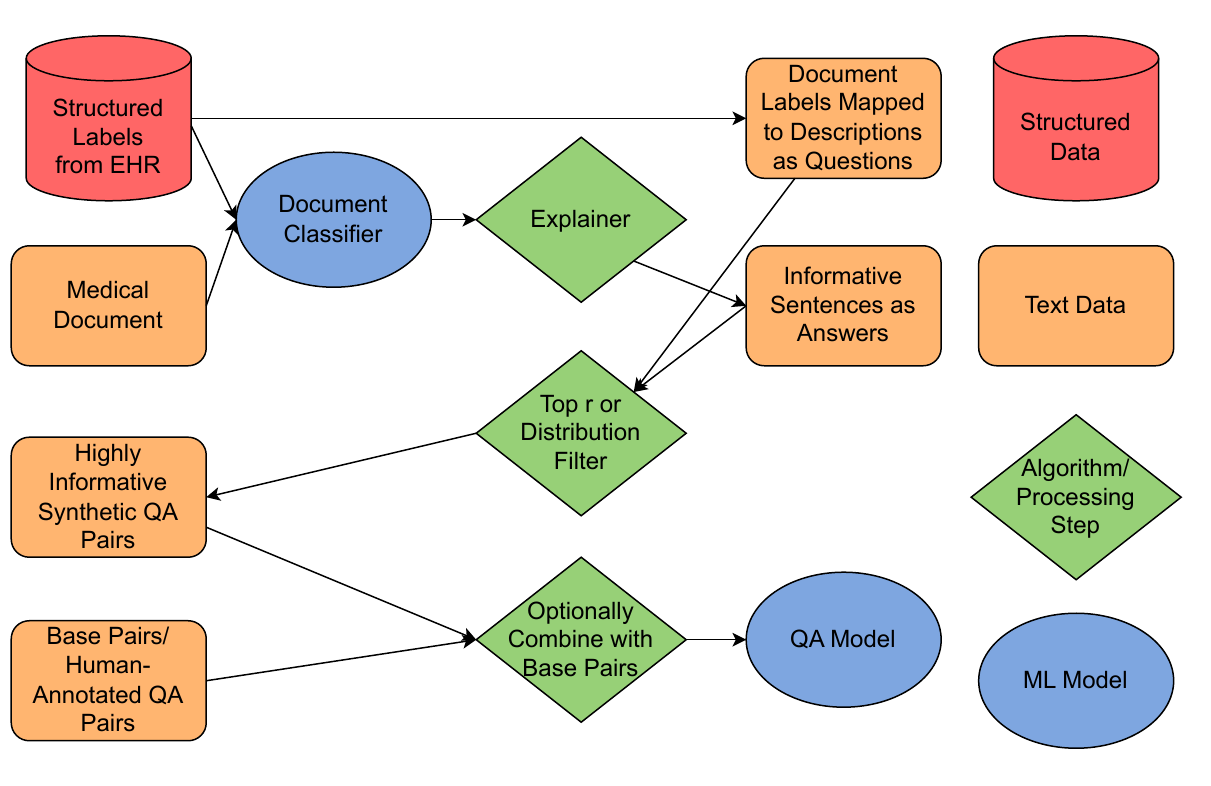} 
  \vspace*{-5mm}
  \caption{XAIQA for generating synthetic QA pairs from medical documents and augmenting the training data or prompts of extractive QA models.  The shapes without arrows indicate shape types in the flow diagram. 
  }
  \label{fig:xaiqa_diagram}
\end{figure*}

Many document-level search and decision support problems can be framed as question answering (QA).
Given a document context and question, an extractive QA model identifies spans of text likely to contain answers.
Applications in healthcare include checking medical guidelines and policies against patient charts,
querying the medical history of a patient,
and identifying cohorts for clinical trials \citep{demner-fushman_what_2009, patrick_ontology_2012}.
However, annotated QA pairs are difficult to obtain, especially for medical text.
Clinical QA models can be trained on generic QA datasets such as SQuAD \citep{rajpurkar_know_2018} then applied to the medical domain,
but performance suffers \citep{soni_radqa_2022}. It is typically possible to pay clinicians to annotate a small set of medical documents
with QA pairs, but these small datasets often focus on one medical speciality.

In addition to training ML models for medical document QA, large language models (LLMs) such as GPT-4 \citep{openai_gpt-4_2023}
can perform clinical evidence extraction using in-context learning (ICL) \citep{agrawal_large_2022}.  
Providing few-shot examples of Q's and A's via ICL can improve the performance of
LLMs and tailor them to specific problem domains, but for each new domain,
different in-context examples are required \citep{dong_survey_2023}.

We focus on generating clinically accurate extractive QA pair examples that 
answer natural Q's about patient conditions of similar form to Q: "Does the patient have $X$ in their medical history?"
A: "The patient has a history of $X$." Systems that can answer these types of Q's can empower clinicians 
and researchers to quickly surface conditions, procedures, and treatments from patient medical histories.

Explainable artificial intelligence (XAI) focuses on explaining ML model predictions.
We propose XAIQA (Figure~\ref{fig:xaiqa_diagram}), 
a method to construct QA pairs from information naturally associated with documents
in the electronic health record (EHR) such as past medical conditions,
drugs, and procedures.  XAIQA pairs can be used to augment existing QA datasets or LLM prompts for ICL.
XAIQA exploits the natural symmetry between the above QA problem formulation (Does $X$ apply?)
and the problem of explaining document classifiers with a text classification explainer.

Three main advantages 
of XAIQA are (1) scale: XAIQA can form QA pairs for 
as many document, medical code pairs as exist in an EHR,
(2) groundedness: the answer spans come from real medical document text
and are not abstractive, and (3) the ability to produce non-keyword matches between Q's and A's.
Our experiments with physician annotators suggest XAIQA generates QA pairs of greater semantic complexity - 
$2.2\times$ more semantic matches and $3.8\times$ more clinical abbreviations - than natural baselines
and provides statistically significant performance improvements to GPT-4 as an extractive QA model
with about 5\% absolute improvement in ROUGE \citep{lin_rouge_2004} on hard questions.




\section{Related Work}



The need for medical QA data to train QA systems for the healthcare setting has spurned extensive research.
\citet{pampari_emrqa_2018-1} repurposed annotations from medical documents in i2b2\footnote{ \url{https://www.i2b2.org/NLP/DataSets/} } to generate Q's
by asking physicians about i2b2 entities and normalizing the Q's into templates, creating the emrQA dataset. 
\citet{yue_cliniqg4qa_2021} found that A's in the relations subset of emrQA better correspond to human
annotation of Q's than do answers to medications Q's.  
For this reason, we used the relations subset for automated evaluation of XAIQA.
They also argued that because the A's in emrQA are frequently incomplete, exact match (EM) is a poor evaluation metric.
Furthermore, they found that most of the A's in both the medications
and relations subsets contain the key phrase in the Q.  This introduces the need to focus on hard Q's. 
\citet{shinoda_can_2021} proposed a method for identifying Q's in QA datasets that have low lexical overlap with the documents. 
We used their approach to create hard subsets of the emrQA relations and mimicQA \citep{yue_cliniqg4qa_2021} datasets.

Another related area of research is question generation \citep{alberti_synthetic_2019, yue_cliniqg4qa_2021, lehman_learning_2022}. 
Our work differs from these approaches in that we focused specifically on constructing grounded answers to questions of the form "does the patient have $X$?"  
One natural way to create QA pairs of this form is with sentence transformers \citep{reimers_sentence-bert_2019},
given their speed, performance, and availability via the sentence-transformers\footnote{ \url{https://github.com/UKPLab/sentence-transformers} } Python library.  
\citet{nguyen_spbertqa_2022} used this approach with BERT \citep{devlin_bert_2019} to generate QA pairs,
and we compared our method to theirs.  Instead of BERT as the sentence encoder, 
we used ClinicalBERT \citet{alsentzer_publicly_2019-1} for its pretraining in the biomedical and clinical domains
and MPNET \citep{song_mpnet_2020} for its top performance
on sentence embedding benchmarks.\footnote{ \url{https://www.sbert.net/docs/pretrained_models.html} }

Several ML systems for the healthcare domain have used medical code descriptions to add information about 
medical concepts and their relationships to ML models \citep{mullenbach_explainable_2018, vu_label_2020, yuan_code_2022}.
We took inspiration from these approaches. 

As a text classification explainer, 
we used the Masked Sampling Procedure (MSP) \citep{stremmel_extend_2022} for
its state-of-the-art performance explaining long medical document classifiers 
(details in Section~\ref{sec:further_discussion_of_related_work}), but other explainers could be used such as the
Sampling and Occlusion algorithm \citep{jin_towards_2020} or Grad-SHAP \citep{lundberg_unified_2017-2}. 
We modeled our evaluation of synthetic QA pairs on the work from \citet{stremmel_extend_2022}.  
While they compared MSP to other XAI methods for text classification,
we used physician evaluation to compare MSP-generated QA pairs to those generated via sentence transformers.

\section{Datasets}


To generate QA pairs, we used discharge summaries from MIMIC-III \citep{johnson_mimic-iii_2016}
and data splits from prior work \citep{mullenbach_explainable_2018} on classifying
International Classification of Disease (version nine) codes (ICD-9s).
Many subsequent papers have focused on the top 50 most common ICD-9s from this dataset
which limits the diversity of conditions used for the classification task. 
We focused instead on ICD-9s occurring in at least 100 of the 47,724 total discharge summaries from \citet{mullenbach_explainable_2018},
resulting in the same splits and 1,124 total medical condition labels.
We call this dataset “MIMIC-III $\geq$ 100.”  We elected to use MIMIC-III over MIMIC-IV \citep{johnson_mimic-iv_2023}
because of the public availability of these data splits.\footnote{ \url{https://github.com/jamesmullenbach/caml-mimic}
}


We used two QA datasets in our ML evaluations. The first is the emrQA relations dataset
consisting of medical document Q's and A's.
We used a random 5\% sample of the test set consisting of 10,835 QA pairs to measure the performance 
of GPT-4 when provided synthetic QA pairs as in-context examples. 
We also fine-tuned Longformer for extractive QA as in \citet{li_clinical-longformer_2022} 
on subsets of the emrQA relations train set with the addition of synthetic QA pairs,
following the recommendation from \citet{yue_cliniqg4qa_2021} (see Section~\ref{sec:further_dataset_description}).
We then evaluated on the full emrQA test set.

The second QA dataset we used to test models with synthetic QA pairs
is mimicQA from \citet{yue_cliniqg4qa_2021}.
This dataset was released by the authors to be used as a test set only, since the dataset is small. 
It contains 1,287 QA pairs, 975 clinician-verified pairs and 312 which are clinician-generated.

\section{Methods}




\subsection{Explainer-Based Data Augmentation}

\SetKwComment{Comment}{/* }{ */}
\begin{algorithm2e}[ht]
\caption{XAIQA: $d$ is a set of $n$ documents, $h$ a set of $l$ labels, $C$ a classifier, and $E$ an explainer. 
We assume the existence of a $hasLabel$ function that checks if a document is positive for a label and a $description$ 
function that returns the textual description of a label.  $A_i$ is the matrix of importance scores for each label $l$ produced by the 
explainer for the $i$th document.}\label{alg:xaiqa_algo}
\DontPrintSemicolon
\KwData{$d$, $h$} 
\KwResult{augPairs}
\SetKwFunction{FSum}{XAIQA}
\SetKwProg{Fn}{Function}{:}{}
\SetKw{KwBy}{by}
\Fn{\FSum{$d$, $h$, $C$, $E$}}{
  augPairs $\gets$ [ ]\;
  \For{$i=1$ \KwTo $n$}{
    $A_i$ $\gets E(C, d_i)$\;
    \For{$j=1$ \KwTo $l$}{
      \If{$hasLabel(d_i, h_j)$}{
        question $\gets description(h_j)$\;
        score $\gets max(A_i[:,j])$\;
        answerIdx $\gets argMax(A_i[:,j])$\;
        answer $\gets s_{answerIdx}$\;
        augPairs.append(question, answer, $d_i$, score)\;
      }
    }
  }
  \KwRet augPairs\;
}
\end{algorithm2e}

We present Algorithm~\ref{alg:xaiqa_algo} based on the intuition that whatever information in a
document is useful for classifying document labels can also be used to form QA pairs,
where the label descriptions form the Q's and the predictive text in the document form the A's. 
Unsupervised approaches for generating such matches typically maximize 
a similarity metric like cosine similarity.  While this can provide some semantic but not lexical matches between Q's and A's,
it has a critical limitation:  Cosine similarity is maximized when the tokens in the Q
are the same as the tokens in the A.  Our approach is not limited in this way, as a classification
algorithm will use whatever text in a document most improves classification performance. 

Algorithm~\ref{alg:xaiqa_algo} takes as input a set of $n$ documents $d = [d_1, d_2,\dots, d_n]$ tokenized into 
$m$ sentences $s = [s_1, s_2,\dots, s_m]$ having up to $l$ labels $h = [h_1, h_2,\dots, h_l]$, a classifier $C$
which outputs label scores for each document, and a sentence explainer $E$.  Note that the number of sentences $m$ 
will vary per document $d_i$ based on the content as will the value of each label (either 1 or 0 for each $h_j$).
$E$ uses $C$ to produce an $m \times l$ matrix $A_i$ for each document $d_i$
by running inference
on perturbed versions of the sentences in $d_i$. $A_i$ contains importance scores for
each sentence, label pair 
and can be used to identify the most important sentences for each label prediction.
More generally, the explainer can use any feature attribution method to assign 
importance scores to sentences using the trained classifier.  

In this way Algorithm~\ref{alg:xaiqa_algo} builds
QA pairs by iterating through documents and labels, collecting the labels as Q's and the
sentences with the highest scores from the explainer as A's along with the sentence scores.  
QA pairs are produced for each 
document, label pair for which the label is positive and can later be filtered by score.
That is, after running Algorithm~\ref{alg:xaiqa_algo}, 
$augPairs$ can be sorted to select the top $r$ QA pairs and associated contexts. 
These can then be used directly for model training, combined with other QA pairs, or used for ICL.

In addition to Algorithm~\ref{alg:xaiqa_algo},
we present a post-processing step (Algorithm~\ref{alg:xaiqa_pp_algo})
based on initial results in Section~\ref{sec:main_phys_eval_res} which uses 
ClinicalBERT to further segment answer sentences.  See Section~\ref{sec:post_processing}
for more information.

\subsubsection{Generating QA Pairs}

To train a classifier to extract medical concepts,
we selected Longformer \citep{beltagy_longformer_2020} for its demonstrated performance on long medical documents
such as discharge summaries \citep{li_clinical-longformer_2022} and fine-tuned it
as described in Section~\ref{sec:generating_synth_qa_pairs}
on the MIMIC-III $\geq$ 100 train set.
See performance in Section~\ref{sec:generating_synth_qa_pairs}. 
We then used MSP, Longformer, and Algorithm~\ref{alg:xaiqa_algo} to 
generate QA pairs from the combined validation and test sets of MIMIC-III $\geq$ 100.
We used the same combined dataset to generate QA pairs with the sentence transformer models. 

\subsection{Evaluating QA Pairs}

We evaluated synthetic QA pairs in an expert evaluation and by using them
as few-shot examples for GPT-4.  We also tested adding these pairs to a training dataset 
used to fine-tune Longformer.
To show our method produces Q's and A's pertaining to complex clinical relationships
and not just keywords, physician annotators counted non-keyword matches and abbreviations
in the A spans in addition to lexical matches.  
In the ML evaluation, we focused on whether synthetic pairs improve LLM performance,
particularly on hard questions. 

As baselines, we tested ClinicalBERT \citep{alsentzer_publicly_2019-1} and MPNET \citep{song_mpnet_2020}
as sentence transformers for embedding sentence, ICD description pairs and selecting the pairs with the highest similarity score. 
For both methods we used the mean token embedding over the CLS token based on results from \citet{reimers_sentence-bert_2019}.
To create synthetic pairs for all methods, we selected samples from among the
top $r=5,914$ pairs with the highest score (cosine similarity or explainer score),
as this is equal to the size of the 1\% subset of emrQA
and enabled us to test 1:1, 2:1, and 5:1 ratios of base to synthetic data in the Longformer evaluation.

\subsubsection{Physician Evaluation}

\begin{table*}[t]
    \centering 
    \caption{Selected examples from physician evaluation.
    In the question, we omit "Does the patient have $X$ in their medical history?"
    Sem (correct answer which is not a string match, abbreviation, or negation),
    Lex (Lexical/string match), Abbr (Abbreviation), Algo (synthetic QA algorithm).}
    \begin{tabular}{p{3cm}p{5cm}p{0.8cm}p{0.8cm}p{0.8cm}p{1.4cm}}
        \toprule
        \textbf{Question} & \textbf{Answer} & \textbf{Sem} & \textbf{Lex} & \textbf{Abbr} & \textbf{Algo}\\
        \midrule
        \midrule
        Unspecified hypothyroidism & Levothyroxine 100 mcg Tablet Sig: One (1) Tablet PO DAILY (Daily). & 1 & 0 & 0 & XAIQA\\
        \midrule
        Esophageal reflux  & Past Medical History: End stage dementia with recurrent aspiration Osteoporosis GERD Social History: Lives in nursing home. & 0 & 0 & 1 & XAIQA\\
        \midrule 
        Unspecified glaucoma & Glaucoma 11. & 0 & 1 & 0 & MPNET\\
        \midrule
        Dysphagia, unspecified & Swallowing difficulties. & 1 & 0 & 0 & MPNET\\
        \midrule 
        Acute osteomyelitis involving ankle and foot & Irrigation and debridement, extensive, to bone of left lower extremity degloving injury and grade 3A open tibia fibula fracture. & 1 & 0 & 0 & CBERT\\
    \bottomrule
    \end{tabular}
    \label{tab:examples} 
\end{table*}

Two physicians worked independently to annotate the same set of 800 blinded QA pairs.
We included: 200 pairs generated by randomly selecting ICD descriptions as Q's and sentences
from the MIMIC-III $\geq$ 100 discharge summaries as A's,
200 sampled from the top $r$ pairs generated from this dataset using XAIQA,
and 400 generated using cosine similarity between ICD descriptions and discharge summary sentences.
Of the 400, 200 came from the top $r$ pairs from ClinicalBERT as the sentence encoder and 200 from MPNET.

The physicians were asked to annotate correct A's to the synthetic Q's about ICD diagnoses, A's
with direct string matches to the key clinical concepts in the question, abbreviation matches between 
QA concepts, and negations. We considered QA pairs with correct A's that were not string matches 
to be interesting semantic matches. Such QA pairs can add diversity to a train dataset or prompt,
providing information about complex relationships between medical concepts.  See 
the exact annotation instructions in Section~\ref{sec:sup_phys_eval_meth}.

To combine results from the two annotators, we counted semantic matches as those 
marked as neither lexical (string/keyword) matches nor abbreviations according to each physician but
marked as correct A's by at least one. 
We counted lexical matches and abbreviations if either physician marked a pair as such.
We used two sample T-Tests without assuming equal variances to compare results from each method.

\begin{table*}[t]
    \centering
    \caption{The number (proportion) of correct answers according to at least one physician 
    but not marked as lexical matches (Lex) or abbreviations (Abbr) by either physician are denoted as semantic matches (Sem).  
    ClinicalBERT is denoted as CBERT.
    200 synthetic QA pairs were provided from each algorithm.
    Algorithms which are significantly better ($\alpha = 0.05$) than the next 
    best algorithm are bolded.}
    \begin{tabular}{llll}
    \toprule
      \textbf{Algo} & \textbf{Sem}  & \textbf{Abbr} & \textbf{Lex} \\
      \midrule
      \midrule
      XAIQA & \textbf{65 (0.325)} & \textbf{45 (0.225)} & 108 (0.540) \\
      CBERT & 30 (0.150) & 12 (0.060) & 128 (0.640) \\
      MPNET & 14 (0.070) & 2 (0.010) & \textbf{181 (0.905)} \\
      Random & 5 (0.025) & 2 (0.010) & 6 (0.030) \\
      \bottomrule
    \end{tabular}
    \label{tab:expert_table_or}
  \end{table*}

\subsubsection{ML Evaluation}

A benefit of our method and the sentence transformers approaches
is that they never show example A's
to the QA model that exceed a sentence in length.
As such, we evaluated all methods primarily on ROUGE-2 (we refer to it here as ROUGE),
which measures bi-gram recall for predicted A spans \citep{lin_rouge_2004}. 
We looked additionally at F1 and EM scores, but because we never showed the model more than one sentence in a
synthetic QA pair, predicted A spans did not trivially maximize ROUGE by containing entire passages
(See predicted A lengths in appendix Table~\ref{tab:pred_ans_lens_emrqa_gpt4_table}). 
Furthermore, by aiming to maximize Rouge, we privileged methods which fully answered each question. 

To measure downstream performance of synthetic QA pairs,
we included them as few-shot examples in GPT-4 prompts and compared with zero-shot inference. 
For each method, we sampled 10 QA pairs from the top $r$ pairs based on the context length of the
8k GPT-4 model when providing 100 characters of context to the left and right of each A span for each in-context example. 
We elected to provide A's with some limited context rather than entire documents based on \citet{liu_lost_2023}
and our initial experiments.
We found that providing the entire document as context per in-context example made performance suffer drastically compared to the 
zero-shot setting (see Table~\ref{tab:100_vs_full}). 
Because example A spans and test documents varied in length, 
we automatically reduced the prompts for GPT-4 one QA pair at a time if
the prompt exceeded the 8k context window.
Prompts and API version are provided in Section~\ref{sec:sup_ml_eval_meth}.

\subsubsection{Hard Subset Creation}

Many of the examples in emrQA relations and other QA datasets are direct string matches. 
We are interested in the ability of a model to extract complete A's to medical Q's when the
Q's and A's have low lexical overlap.
At runtime, one cannot know which method to select based on lexical overlap between Q's and A's,
so we instead computed query context overlap (QCLO) as in \citet{shinoda_can_2021-1}. 
See the equation for QCLO in Section~\ref{sec:hard_subset}. 

\section{Results}

\subsection{Physician Evaluation}
\label{sec:main_phys_eval_res}

As shown in Table~\ref{tab:expert_table_or}, XAIQA produced 
significantly more semantic matches than the baselines ($2.2\times$ more than ClinicalBERT with $p<0.001$).
Our method also identified significantly more abbreviations than both baselines ($3.8\times$ more than ClinicalBERT with $p<0.001$).
MPNET found the most lexical matches.
We omitted negations from Table~\ref{tab:expert_table_or}, as each method identified 10 or fewer, limiting statistical power.
Section~\ref{sec:sup_phys_eval_res} contains results from individual physicians, 
examples of false positive QA pairs with physician commentary, 
and motivation for Algorithm~\ref{alg:xaiqa_pp_algo} to post-process XAIQA
pairs before providing them to QA models. Annotator agreement is reported in Table~\ref{tab:agreement}.

Table~\ref{tab:examples} depicts examples selected by the physicians 
following the blind experiment to illustrate some relationships represented in the synthetic QA pairs.  
For example, XAIQA recognized the semantic match between the use of a medication (levothyroxine), 
and the presence of “unspecified hypothyroidism.”  This drug is prescribed for virtually no other condition. 
XAIQA also captured many abbreviations, as in “esophageal reflux,” where the abbreviation “GERD” is semantically equivalent
(the acronym stands for “gastroesophageal reflux disease”). 
MPNET found the lexical match “unspecified glaucoma” and “glaucoma.”
String matches of this kind were typical for MPNET, though we note MPNET is capable of identifying some semantic matches,
e.g., “dysphagia” is difficulty swallowing.
ClinicalBERT recognized that if “irrigation and debridement, extensive, to bone” are required, this is very
likely to reflect “osteomyelitis (bone infection), involving ankle and foot.”

\subsection{ML Evaluation}

\begin{table*}[h]
    \centering 
    \caption{Test set results for GPT-4 with synthetic QA pairs.  XAIQA with ClinicalBERT (CBERT) used for post-processing denoted as PP.
    Exact match denoted as EM.}
    \begin{tabular}{p{1.6cm}p{2.6cm}p{3.25cm}p{3.25cm}p{3.25cm}}
      \toprule
      \textbf{Dataset} & \textbf{Method} &                 \textbf{ROUGE} &                    \textbf{F1} &                    \textbf{EM} \\ \hline \hline
      \multirow{5}{*}{emrQA} & Zero-Shot & 0.667 [0.659 - 0.675] & 0.737 [0.730 - 0.743] & 0.491 [0.482 - 0.501] \\
 & CBERT & 0.697 [0.690 - 0.705] & 0.744 [0.737 - 0.750] & 0.514 [0.505 - 0.524] \\
 & MPNET & 0.696 [0.688 - 0.703] & 0.747 [0.740 - 0.753] & 0.518 [0.509 - 0.527] \\
 & XAIQA Raw & 0.703 [0.695 - 0.711] & 0.735 [0.728 - 0.741] & 0.509 [0.500 - 0.518] \\
 & XAIQA PP & 0.708 [0.701 - 0.716] & 0.746 [0.739 - 0.752] & 0.518 [0.509 - 0.527] \\ \hline \hline
 \multirow{5}{*}{mimicQA} & Zero-Shot & 0.520 [0.497 - 0.543] & 0.600 [0.579 - 0.621] & 0.300 [0.275 - 0.326] \\
 & CBERT & 0.550 [0.528 - 0.573] & 0.606 [0.586 - 0.627] & 0.314 [0.289 - 0.340] \\
 & MPNET & 0.556 [0.532 - 0.580] & 0.613 [0.593 - 0.634] & 0.324 [0.298 - 0.349] \\
 & XAIQA Raw & 0.555 [0.531 - 0.579] & 0.598 [0.577 - 0.620] & 0.316 [0.290 - 0.342] \\
 & XAIQA PP & 0.562 [0.538 - 0.586] & 0.613 [0.593 - 0.633] & 0.309 [0.285 - 0.333] \\ 
\bottomrule
\end{tabular}
          \label{tab:gpt4_table}
      \end{table*}

\begin{figure}[h]
  \centering
  \includegraphics[width=3in]{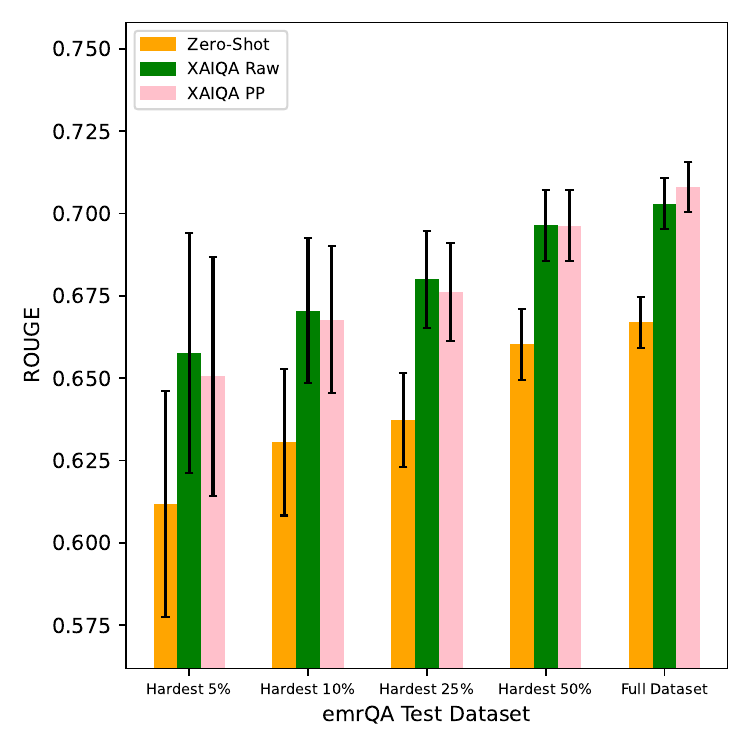}
  \vspace*{-5mm}
  \caption{ROUGE scores of GPT-4 with XAIQA pairs on emrQA relations. PP denotes post-processing.}
  \label{fig:gpt4_emrqa_ROUGE_zero}
\end{figure}

\begin{figure}[h]
  \centering
  \includegraphics[width=3in]{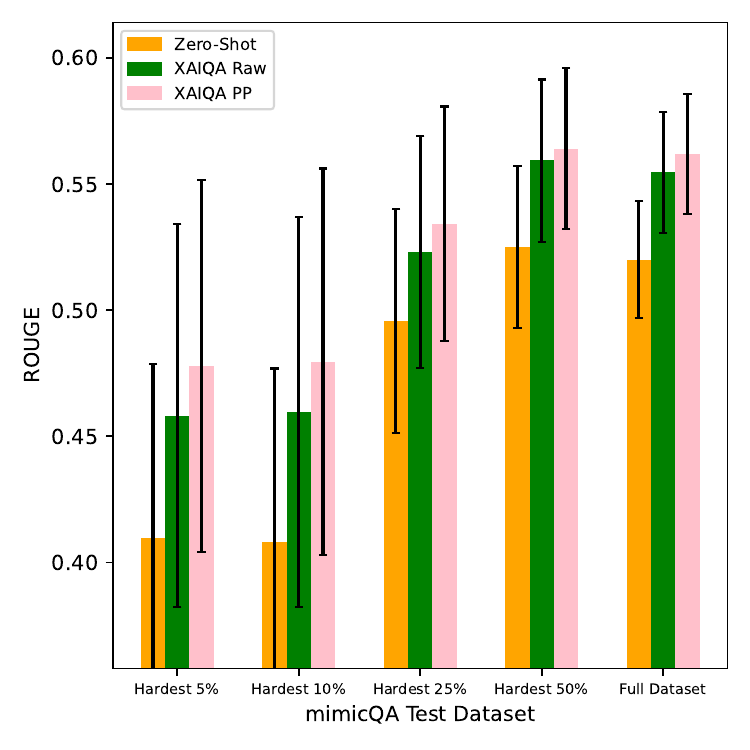}
  \vspace*{-5mm}
  \caption{ROUGE scores of GPT-4 with XAIQA pairs on mimicQA. PP denotes post-processing.}
  \label{fig:gpt4_mimicqa_ROUGE_zero}
\end{figure}

\begin{table}[h]
    \centering 
    \caption{Mean test ROUGE on hardest 5 and 10\% of Q's.
    }
    \begin{tabular}{p{1.6cm}p{2.6cm}p{0.8cm}p{0.8cm}}
      \toprule
      \textbf{Dataset} & \textbf{Method} & \textbf{5\%} & \textbf{10\%} \\ \hline \hline
      \multirow{5}{*}{emrQA} & Zero-Shot & 0.612 & 0.630 \\ 
 & CBERT & 0.646 & 0.663 \\
 & MPNET & 0.634 & 0.650 \\
 & XAIQA Raw & 0.658 & 0.670 \\
 & XAIQA PP & 0.650 & 0.668 \\ \hline \hline
 \multirow{5}{*}{mimicQA} & Zero-Shot & 0.410 & 0.408 \\ 
 & CBERT & 0.440 & 0.440 \\
 & MPNET & 0.443 & 0.443 \\
 & XAIQA Raw & 0.458 & 0.460 \\
 & XAIQA PP & 0.478 & 0.480 \\ \hline \hline
\bottomrule
\end{tabular}
          \label{tab:gpt4_hardest_table}
      \end{table}

Figure~\ref{fig:gpt4_emrqa_ROUGE_zero} compares adding in-context examples from XAIQA to
zero-shot performance of GPT-4 on 10,835 emrQA relations test pairs as well as the hardest 5, 10, 25, and 50\%
in terms of QCLO.  Figure~\ref{fig:gpt4_mimicqa_ROUGE_zero} shows the same for the mimicQA test set (1,287 total pairs). 
Performance is reported as the mean of the performance metric over 1,000 bootstrap samples drawn from the test set with 95\% 
confidence intervals. 

XAIQA provides statistically significant performance improvements (confidence intervals do not overlap) 
over zero-shot GPT-4 on the full emrQA relations test set as well as the hardest 25 and 50\% of examples.
Test set size decreases when subsetting to difficult examples, widening the confidence intervals,
but the pattern in the mean ROUGE score over bootstrap iterations is clear on the hardest subsets of emrQA and mimicQA, 
where the performance improvement is about 5\%.

As shown in Table~\ref{tab:gpt4_table}, XAIQA with post-processing outperforms sentence transformers approaches
in terms of ROUGE with the same F1 scores.
Comparing raw and post-processed XAIQA, we observe post-processing improves F1.
MPNET performance is strong in F1 and EM on the full datasets,
consistent with the results of the expert evaluation
where MPNET found the most correct A spans, despite most of these being easy examples.  
On the hardest 5 and 10\% of Q's, XAIQA outperforms the sentence transformers approaches
in terms of ROUGE (Table~\ref{tab:gpt4_hardest_table}).
Section~\ref{sec:sup_results_all} includes ROUGE and F1 plots for all hard subsets tested.

Fine-tuning Longformer on different ratios of base to synthetic QA pairs
(appendix Figure~\ref{fig:longformer_emrqa}) improves performance once the ratio is 2:1 base to synthetic. 
There, MPNET offers significant gains.
At a 5:1 ratio, XAIQA with post-processing and ClinicalBERT outperform no synthetic data.
We hypothesize Longformer needs more base data to benefit from semantic relationships present in the QA pairs from
XAIQA. At all ratios, Longformer outperforms GPT-4, underscoring the value of fine-tuning when data is abundant.




\section{Discussion}

We presented a novel method, XAIQA, to generate QA pairs that can inform models of complex relationships between medical document Q's and A's.
With XAIQA, ML practitioners can generate targeted examples at scale 
for any medical concept to add to an LLM prompt without the need for human annotation.
Our approach excels at introducing semantic and clinical abbreviation relationships between Q's and A's with
$2.2\times$ more semantic matches and $3.8\times$ more clinical abbreviations compared to ClinicalBERT.  
XAIQA significantly improves GPT-4 as an extractive QA model with up to 5\% improvement on hard questions.

Future work is required to analyze optimal base to synthetic ratios for both LLM ICL and LM fine-tuning
to help ML practitioners understand how to make the most of synthetic QA data.
Beyond our experiments presented here, we believe XAIQA can turn any document-level classification dataset into an extractive QA dataset, 
augmenting existing QA data or LLM prompts with grounded and semantically complex answer spans.

\clearpage
\bibliography{stremmel22}

\clearpage
\appendix

\section{Further Discussion of Related Work}
\label{sec:further_discussion_of_related_work}

\citet{yue_cliniqg4qa_2021} benchmarked BERT models for medical document QA, 
and \citet{li_clinical-longformer_2022} trained sparse-attention LMs for long document QA. We used their approach
to fine-tune Longformer \citep{beltagy_longformer_2020}. 

We selected MSP for its superior performance to the Sampling and Occlusion algorithm \citep{jin_towards_2020},
which demonstrated better performance than Grad-SHAP \citep{lundberg_unified_2017-2} for explaining text classifiers.
Additionally, \citep{stremmel_extend_2022} demonstrate the runtime efficiency of MSP in the context of long medical documents
with fine-tuned sparse-attention LMs. 

The core idea of the algorithm is that MSP recovers the text blocks which are predictive of document labels by randomly masking 
a subset of text blocks over many iterations and then computing the difference in the probabilities of each label from the classifier
when each text block is unmasked versus masked.  These differences are averaged over all masking iterations.  Because text blocks 
are masked jointly, MSP simulates cases where mulitple pieces of information pertaining to document labels are included and excluded.
This joint masking approach also leads to runtime efficiencies over simply masking each text block.  We used sentence-level text blocks 
instead of fixed length blocks to provide more natural answer spans to our synthetic questions.

\section{Additional Description of Datasets}
\label{sec:further_dataset_description}

\citet{yue_cliniqg4qa_2021} found that only 5\% of the total relations train set
is necessary to obtain the same level of performance as training on the full dataset.
We trained Longformer on up to 5\% of the dataset for this reason.
We used the full validation set for early stopping after five epochs of no improvement.
We evaluated the Longformer models on the full emrQA relations test set but randomly 
sampled the test set to evaluate the GPT-4 models given the high cost of using the OpenAI API.

\section{Supplemental Methods}
\label{sec:supplemental_methods}

\subsection{Explainer-Based Data Augmentation}
\label{sec:sup_ebda}

\subsection{Post-Processing}
\label{sec:post_processing}

We used Algorithm~\ref{alg:xaiqa_pp_algo} to produce more concise A's by further splitting sentences identified by the explainer
and selecting the segment with the highest cosine similarity to the medical concept question, 
balancing the idea of classification-based and similarity-based A selection.  
Specifically, Algorithm~\ref{alg:xaiqa_pp_algo}
uses the explainer to identify predictive sentences and cosine similarity to resolve long sentences
with bullet points, numbered lists, or semi-colons, as are often found in medical documents.
When applying post-processing in our experiments, we used ClinicalBERT to generate question and segment embeddings.

Given $augPairs$ from Algorithm~\ref{alg:xaiqa_algo} and a PunktSentenceTokenizer\footnote{ \url{https://www.nltk.org/api/nltk.tokenize.PunktSentenceTokenizer.html} }
from the natural language toolkit (NLTK) Python library as $punkSentTok$, where the tokenizer is initialized with the characters 
\verb-'.', '?', '!', '•', '\-', ';', '0)', '1)', '2)', '3)', '4)', '5)', '6)', '7)', '8)', '9)',
Algorithm~\ref{alg:xaiqa_pp_algo} iterates through a list of $r$ $augPairs$ and extracts the Q's, A's, contexts, and scores.  Answer sentences are 
split into shorter segments using $punkSentTok$ to create segments on bullet points, numbered lists, and semi-colons, as are often found in medical documents.
These segments are then embedded using a sentence embedding function $sentEmb$ which applies one of the sentence encoders with mean pooling of the output tokens.
Question and answer segment embeddings are compared and the segment with the highest cosine similarity to the question is selected and used in an updated 
version of $augPairs$ which we refer to as $augPairsPP$, the output of Algorithm~\ref{alg:xaiqa_pp_algo}.

\SetKwComment{Comment}{/* }{ */}
\begin{algorithm2e}[ht]
  \caption{XAIQA post-processing: $augPairs$ is a list of $r$ (question, answer, context, score) tuples produced by XAIQA.  $punkSentTok$ is a 
  sentence tokenizer.  We assume the existence of a sentence embedding function $sentEmb$ and a function $cosSim$ that compute cosine similarity
  between embeddings.}\label{alg:xaiqa_pp_algo}
  \DontPrintSemicolon
  \KwData{augPairs, punkSentTok} 
  \KwResult{augPairsPP}
  \SetKwFunction{FSum}{XAIQAPP}
  \SetKwProg{Fn}{Function}{:}{}
  \SetKw{KwBy}{by}
  \Fn{\FSum{augPairs, punkSentTok}}{
    augPairsPP $\gets$ [ ]\;
    \For{$i=0$ \KwTo $r$}{
      question $\gets augPairs[i][0]$\;
      answer $\gets augPairs[i][1]$\;
      context $\gets augPairs[i][2]$\;
      score $\gets augPairs[i][3]$\;
      splitAnswers $\gets punkSentTok.tokenize(answer)$\;
      aEmb $\gets sentEmb(splitAnswers)$\;
      qEmb $\gets sentEmb(question)$\;
      topMatch $\gets argMax(cosSim(aEmb, qEmb))$\;
      answerPP $\gets splitAnswers[topMatch]$\;
      augPairsPP.append(question, answerPP, context, score)\;
    }
    \KwRet augPairsPP\;
  }
  \end{algorithm2e}

The sentence splitting heuristics of the post-processing algorithm could be applied before training the classifier;
however, for simplicity, we used the NLTK sentence tokenizer \citep{loper_nltk_2002} to
generate sentences for all methods, which, by default,
does not split on bulleted or numbered lists.  

\subsection{Generating QA Pairs}
\label{sec:generating_synth_qa_pairs}

\subsubsection{Fine-Tuning Longformer for Classification}

We fine-tuned Longformer with a 4,096 token maximum sequence length for multi-label classification using 
PyTorch \citep{paszke_pytorch_2019} (version 1.12.1) and transformers version 4.22.1 \citep{wolf_transformers_2020}.

We used an effective batch size of 128 
(two samples per GPU with eight gradient accumulation steps on an Azure Standard\_ND40rs\_v2 machine with 32 GB GPU RAM per GPU and eight GPUs in total)
and the AdamW \citep{loshchilov_decoupled_2019} optimizer with 0.01 weight decay, 500 warm-up steps and early stopping after five evaluations 
(occurring every 50 steps) with no improvement in validation loss.

\subsubsection{Performance of Longformer Classifier}

\begin{table*}[h]
    \centering
    \caption{Performance of LMs fine-tuned for multi-label classification on MIMIC-50.}
    \begin{tabular}{p{5.7cm}p{3.7cm}p{3.7cm}}
    \toprule
    \textbf{Model} & \textbf{Micro-Average Precision} & \textbf{Macro-Average Precision} \\ 
    \midrule
        BigBird Base (Raw Text) & 0.584 [0.572 - 0.595] & 0.4763 [0.466 - 0.487] \\ 
        BigBird Base (Clean Text) & 0.659 [0.648 - 0.670] & 0.5526 [0.541 - 0.564] \\ 
        BigBird Large (Raw Text) & 0.629 [0.618 - 0.641] & 0.5249 [0.514 - 0.536] \\ 
        BigBird Large (Clean Text) & 0.698 [0.687 - 0.708] & 0.6003 [0.590 - 0.611] \\ 
        Longformer Base (Raw Text) & 0.640 [0.629 - 0.652] & 0.5342 [0.524 - 0.545] \\ 
        Longformer Base (Clean Text) & 0.652 [0.641 - 0.664] & 0.5529 [0.542 - 0.563] \\ 
    \bottomrule
    \end{tabular}
    \label{tab:mimic50clf}
\end{table*}

We compared several classifiers to use for the MSP explainer on the MIMIC-50 dataset from \citet{mullenbach_explainable_2018},
including base and large versions of Longformer and Big Bird \citep{zaheer_big_2021-2}
with and without cleaning the text of the discharge summaries using the logic from \citet{stremmel_extend_2022}.  
Performance on the 1,730 test documents from MIMIC 50 are reported in Table~\ref{tab:mimic50clf}.
All pretrained model checkpoints were taken from the Hugging Face model hub \footnote{ \url{https://huggingface.co/models} }.

Based on the MIMIC-50 experiments, we selected the Longformer base classifier 
for it's performance without text cleaning, enabling easier downstream application of the model. 
We fine-tuned Longformer base on “MIMIC-III $\geq$ 100” resulting in the performance in Table~\ref{tab:mimicgeq100clf}.

\begin{table*}[h]
    \centering
    \caption{Performance of Longformer base for multi-label classification on MIMIC-III $\geq$ 100.}
    \begin{tabular}{p{5.7cm}p{3.7cm}p{3.7cm}}
    \toprule
    \textbf{Model} &  \textbf{Micro-Average Precision} & \textbf{Macro-Average Precision} \\
    \midrule
        Longformer Base (Raw Text) & 0.461 [0.4553- 0.467] & 0.221 [0.216 - 0.226] \\
    \bottomrule
    \end{tabular}
    \label{tab:mimicgeq100clf}
\end{table*}

We expect that higher performing multi-label classifiers would yield better synthetic QA pairs using XAIQA and leave
these experiments for future work.

\subsection{Physician Evaluation}
\label{sec:sup_phys_eval_meth}

The physicians were asked to annotate the following four attributes for each example:

\begin{enumerate}
  \item Correct answers to the synthetic questions about ICD diagnoses where the question is whether
the patient has the medical condition:
This required the correct form of a disease.  For example, if a specific disease form was posed in the question field,
that form should be supported in the answer field.  If the question was for a general form, then a specific form in the answer
field which implied the general form or a general answer would be sufficient.
  \item Direct string matches between clinical concepts in the questions and clinical concepts in the answers:
This field was marked if the example was a correct answer and the evidence included the same key clinical words as the diagnosis. 
Casing, misspelling, varying forms of a word (e.g.,'ing'), and punctuation did not matter. 
Anything very close to a string match was considered a string match,
though the exact interpretation of close was left to each physician's
judgment.  The idea was to record "easy" matches.
  \item Abbreviation matches between question and answer concepts: This field was marked if the example
  was a correct answer and the key clinical
  words in the answer text were an abbreviated form of the diagnosis.
  \item Negations: This field was marked if the example was a correct answer and the diagnosis was refuted by the provided text.
\end{enumerate}

\subsection{ML Evaluation}
\label{sec:sup_ml_eval_meth}

\subsubsection{Performance with Entire Documents In-Context}
\label{sec:100_vs_full}

\begin{table*}[h]
    \centering 
    \caption{emrQA test set results for GPT-4 with synthetic QA pairs from MPNET and XAIQA.  
    We tested both approaches using the full document as context with 5 in-context examples and
    a window of 100 characters to the left and right of the answer as context with 10 in-context examples.}
    \begin{tabular}{p{2.6cm}p{1.4cm}p{3.25cm}p{3.25cm}p{3.25cm}}
      \toprule
      Method &  Context       &          ROUGE &                    F1 &                    EM \\
\midrule
MPNET & Full & 0.483 [0.465 - 0.501] & 0.581 [0.564 - 0.597] & 0.309 [0.290 - 0.328] \\
XAIQA Raw & Full & 0.515 [0.497 - 0.534] & 0.583 [0.565 - 0.600] & 0.347 [0.327 - 0.367] \\
MPNET & Window & 0.696 [0.688 - 0.703] & 0.747 [0.740 - 0.753] & 0.518 [0.509 - 0.527] \\
XAIQA Raw & Window & 0.703 [0.695 - 0.711] & 0.735 [0.728 - 0.741] & 0.509 [0.500 - 0.518] \\
\bottomrule
\end{tabular}
          \label{tab:100_vs_full}
      \end{table*}

\subsubsection{GPT-4 Prompts}
\label{sec:gpt4_prompts}

\begin{table*}[t]
    \centering 
    \caption{Prompts for GPT-4}
    \begin{tabular}{p{1.2cm}p{14cm}}
        \textbf{Setting} & \textbf{Prompt}\\
        \midrule
        \midrule
        Zero-Shot & Extract a span of text from the medical document to answer the supplied medical question with evidence from the document.
        \newline
        \newline
        For example, if you are given the following basic example:
        \newline
        \newline
        Question: "Does the patient have hypertension"
        \newline
        Document: "John Smith is a 56 year-old male. He was diagnosed with hypertension on 04-04-22. John is at risk for stroke."
        \newline
        \newline
        You should answer with:
        \newline
        \newline
        \{"start\_idx": 35, "span\_text": "He was diagnosed with hypertension on 04-04-22."\}
        \newline
        \newline
        Answer questions using only text found in the document. REMEMBER: Return only properly formatted JSON. Do not return any additional text. You are an expert in producing perfectly formatted JSON responses and no additional text. You always remember to close bracks when outputting JSON. You are an expert in medicine and medical document review. You provide perfect answers to questions about patients by carefully reading their medical documents and using your clinical knowledge.
        \newline
        \\
        \midrule
        Few-Shot & Extract a span of text from the medical document to answer the supplied medical question with evidence from the document.
        \newline
        \newline
        For example, if you are given the following basic example:
        \newline
        \newline
        Question: "Does the patient have hypertension"
        \newline
        Document: "John Smith is a 56 year-old male. He was diagnosed with hypertension on 04-04-22. John is at risk for stroke."
        \newline
        \newline
        You should answer with:
        \newline
        \newline
        \{"start\_idx": 35, "span\_text": "He was diagnosed with hypertension on 04-04-22."\}
        \newline
        Answer questions using only text found in the document. REMEMBER: Return only properly formatted JSON. Do not return any additional text. You are an expert in producing perfectly formatted JSON responses and no additional text. You are an expert in medicine and medical document review. You provide perfect answers to questions about patients by carefully reading their medical documents and using your clinical knowledge.
        \newline
        \newline
        Here are some additional examples:
        \newline
        ...
    \end{tabular}
    \label{tab:prompts} 
\end{table*}

Prompts for GPT-4 were as shown in Table~\ref{tab:prompts}. 

For the few-shot prompts, after including examples, we provided the suffix:
"Now that you've seen some examples of what to do, extract a span of text from the medical document to answer the supplied medical question with evidence from the medical document. Answer questions using only text found in the document.  REMEMBER: Return only properly formatted JSON. Do not return any additional text.  You are an expert in producing perfectly formatted JSON responses and no additional text. You always remember to close brackets when outputting JSON. You are an expert in medicine and medical document review. You provide perfect answers to questions about patients by carefully reading their medical documents and using your clinical knowledge."

We found that GPT-4 tended to "forget" the original instructions when providing few-shot examples if not reminded at the end using a suffix like this.

For all experiments with GPT-4, we used the 03-15-2023 version of the OpenAI API, the 8k context version of GPT-4, and temperature 0 to encourage extractive answers.

\subsubsection{Longformer Fine-Tuning for QA}
\label{sec:longformer_fine_tuning_for_qa}

We fine-tuned Longformer base for question answering using the code and instructions from \citet{li_clinical-longformer_2022}
documented here \footnote{ \url{https://github.com/luoyuanlab/Clinical-Longformer/tree/main/Question\%20Answering} }.
We used PyTorch \citep{paszke_pytorch_2019} (version 1.12.1) and transformers version 4.21.1 \citep{wolf_transformers_2020}
and fine-tuned at a maximum sequence length of 4,096 tokens with an effective batch size of 128 
(two samples per GPU with eight gradient accumulation steps on an Azure Standard\_ND40rs\_v2 machine with 32 GB GPU RAM per GPU and eight GPUs in total)
and the AdamW \citep{loshchilov_decoupled_2019} optimizer with 0.01 weight decay, 10,000 warm-up steps and early stopping after 10 evaluations 
(occurring every epoch) with no improvement in validation F1.

\subsection{Hard Subset Creation}
\label{sec:hard_subset}

\citet{shinoda_can_2021-1} define query context overlap as $QCLO = |Q \cap C| / |Q|$. 
Here, Q denotes the set of words in the query, C, the set of words in the context,
and QCLO the ratio of intersecting words in Q and C to the number of words in Q. 
Because this metric counts stopwords and question words which may appear in the context as overlapping and
does not count near matches for the same word with a different suffix, e.g., "ing", we
removed stopwords and question words and applied the NLTK
PorterStemmer\footnote{ \url{https://tartarus.org/martin/PorterStemmer/} } to the set of words in Q and C before computing QCLO.

\section{Supplemental Results}
\label{sec:supplemental_results}

\subsection{Physician Evaluation}
\label{sec:sup_phys_eval_res}

\subsubsection(Individual Annotator Results)

\begin{table*}[t]
    \centering
    \caption{Number (proportion) of semantic but not lexical matches or abbreviations (Sem), abbreviations (Abbr),
    and lexical matches (Lex) identified in the expert evaluation.  ClinicalBERT denoted as CBERT.
    200 synthetic QA pairs were provided from each algorithm. Algorithms which are significantly better ($\alpha = 0.05$) than the next 
    best algorithm according to two-sample T-tests without assuming equal variances are bolded.}
    \begin{tabular}{lllllll}
    \toprule
      & \multicolumn{3}{l}{\textbf{Reviewer 1}} & \multicolumn{3}{l}{\textbf{Reviewer 2}} \\
      \midrule 
      \textbf{Algo} &  \textbf{Sem}  & \textbf{Abbr} & \textbf{Lex}  &  \textbf{Sem}  & \textbf{Abbr} & \textbf{Lex} \\
      \midrule
      \midrule
      \multicolumn{1}{l|}{XAIQA} & \textbf{65 (0.325)} & \textbf{44 (0.220)} & \multicolumn{1}{l|}{102 (0.510)} & 79 (0.395) & \textbf{37 (0.185)} & 74 (0.370)\\
      \multicolumn{1}{l|}{CBERT} & 25 (0.630) & 8 (0.040) & \multicolumn{1}{l|}{126 (0.310)} & 73 (0.365) & 11 (0.055) & 62 (0.310)\\
      \multicolumn{1}{l|}{MPNET} & 14 (0.070) & 1 (0.005) & \multicolumn{1}{l|}{\textbf{178 (0.890)}} & 53 (0.265) & 1 (0.005) & \textbf{129 (0.645)}\\
      \multicolumn{1}{l|}{Random} & 3 (0.015) & 0 (0.000) & \multicolumn{1}{l|}{5 (0.025)} & 6 (0.030) & 2 (0.010) & 3 (0.015)\\
      \bottomrule
    \end{tabular}
    \label{tab:expert_table}
  \end{table*}

As shown in Table~\ref{tab:expert_table}, XAIQA produced
more semantic matches than the baselines according to both reviewers, including 
significantly more than the second best method according to reviewer 1 with $2.6\times$ more and ($P < 0.001$).
Our method also identified significantly more abbreviations than both baselines according to both reviewers,
about $5.5\times$ more than the second best method, ClinicalBERT, according to reviewer 1 and $3.4\times$ more according to reviewer 2.
MPNET found the most lexical matches according to both reviewers.

\subsubsection{Annotator Agreement}

\begin{table}[h]
    \centering 
    \caption{Overall agreement and Cohen's Kappa for each field annotated.}
    \begin{tabular}{p{2.5cm}p{2cm}p{2cm}}
      \toprule
      \textbf{Field} & \textbf{Agreement} & \textbf{Cohen's Kappa} \\ 
      Correct Answer & 0.904 & 0.796 \\ 
 String Match & 0.791 & 0.586 \\
 Abbreviation & 0.978 & 0.815 \\
 Negation & 0.984 & 0.228 \\
\bottomrule
\end{tabular}
          \label{tab:agreement}
      \end{table}

\subsubsection{Example Synthetic QA Pairs}



\subsubsection{Motivation for Explainer Post-Processing}

XAIQA approach was more likely to recover sections from the discharge summaries 
such as History of Present Illness and Past Medical History.  For example, the past medical history section 
is recovered in about 34\% of the top synthetic QA pairs generated by the method.  Cosine similarity,
on the other hand, only does this about 7\% of the time with ClinicalBERT and less than 1\% of the time 
with MPNET.  We hypothesize this is because cosine similarity is reduced when sentences contain several clinical concepts.
While recovering this section when it contains the correct A is ideal when presenting information to a human -
it provides useful context - and not a problem for expert evaluation, this finding motivates Algorithm~\ref{alg:xaiqa_pp_algo}
for post-processing.

\subsubsection{Example False Positive Synthetic Pairs}

\begin{table*}[t]
    \centering 
    \caption{Selected false positive examples from physician evaluation.
    In the question, we omit "Does the patient have $X$ in their medical history?"
    Sem (correct answer which is not a string match, abbreviation, or negation),
    Lex (Lexical/string match), Abbr (Abbreviation), Algo (synthetic QA algorithm).}
    \begin{tabular}{p{5cm}p{5cm}p{0.8cm}p{0.8cm}p{0.8cm}p{1.4cm}}
        \toprule
        \textbf{Question} & \textbf{Answer} & \textbf{Sem} & \textbf{Lex} & \textbf{Abbr} & \textbf{Algo}\\
        \midrule
        \midrule
        Venous catheterization, not elsewhere classified & He did develop ARDS and a ventilator associated pneumonia for which Vancomycin, Zosyn and Ciprofloxacin were started & 0 & 0 & 0 & XAIQA\\
        \midrule
        Pneumonia due to escherichia coli [E. coli] & Klebsiella Pneumoniae. & 0 & 0 & 0 & CBERT\\
    \bottomrule
    \end{tabular}
    \label{tab:examples_fp} 
\end{table*}

Despite being pretrained on clinical text, ClinicalBERT made no shortage of mistakes (49 in total out of 200 for annotator 1 
and 65 in total for annotator 2).
In Table~\ref{tab:examples_fp}, we show how ClinicalBERT incorrectly associated “… pneumonia due to Escherichia coli”
with “Klebsiella pneumoniae,”
the genus and species of a different bacterium whose species designator (“pneumoniae”) misled the model into a pneumonia diagnosis,
likely as a result of overlapping subtokens when representing “pneumoniae” and pneumonia.  
Like the other approaches, XAIQA produced some false positive predictions (33 in total out of 200 for annotator 1 
and 47 in total for annotator 2), as in 
the question of "venous catheterization," where the model incorrectly provided an answer related to ARDS,
a bad respiratory failure episode requiring a ventilator and blood draws or indwelling IV lines.
This correlation could account.

\subsection{ML Evaluation}
\label{sec:sup_ml_eval_res}

\subsubsection{Predicted Answer Lengths}

\begin{table}[t]
    \caption{5th, 50th and 95th percentile of predicted answer lengths (in tokens) for GPT-4 on emrQA relations.}
    \resizebox{\columnwidth}{!}{
    \begin{tabular}{llll}
        \toprule
    \textbf{Method} & \textbf{5th} & \textbf{50th} & \textbf{95th} \\
    \midrule
        Zero-Shot & 2 & 8 & 18 \\ 
        CBERT & 2 & 10 & 22 \\ 
        MPNET & 2 & 9 & 19 \\ 
        XAIQA Raw & 2 & 10 & 24 \\ 
        XAIQA PP & 3 & 10 & 22 \\
        \bottomrule
    \end{tabular}
    \label{tab:pred_ans_lens_emrqa_gpt4_table}
    }
\end{table}

\subsubsection{All Methods on Hard Subsets}
\label{sec:sup_results_all}

\begin{figure}[h]
  \centering
  \includegraphics[width=3in]{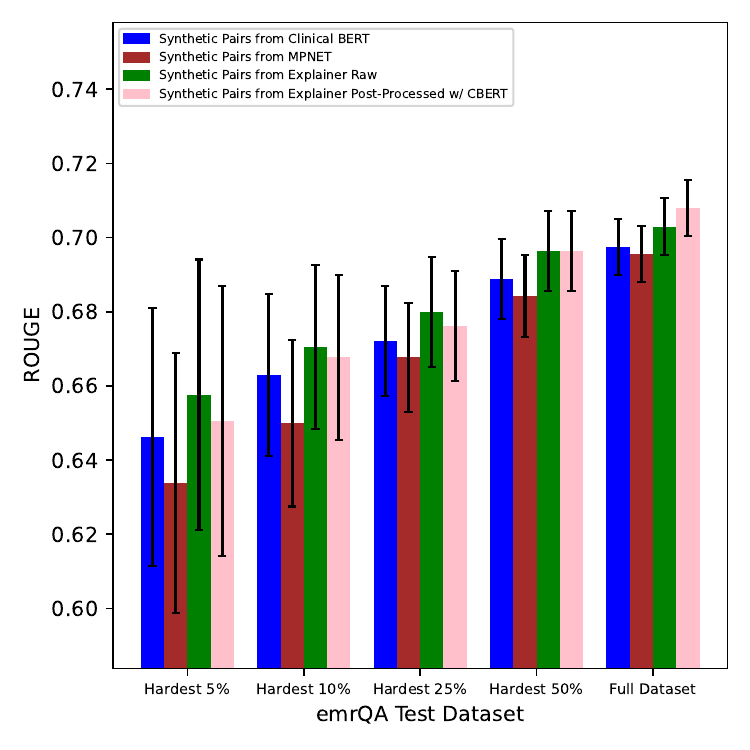}
  \vspace*{-5mm}
  \caption{ROUGE scores of GPT-4 for extractive QA using different synthetic QA pairs on hard subsets of emrQA relations. CBERT is ClinicalBERT.}
  \label{fig:gpt4_emrqa_ROUGE_base}
\end{figure}

\begin{figure}[h]
  \centering
  \includegraphics[width=3in]{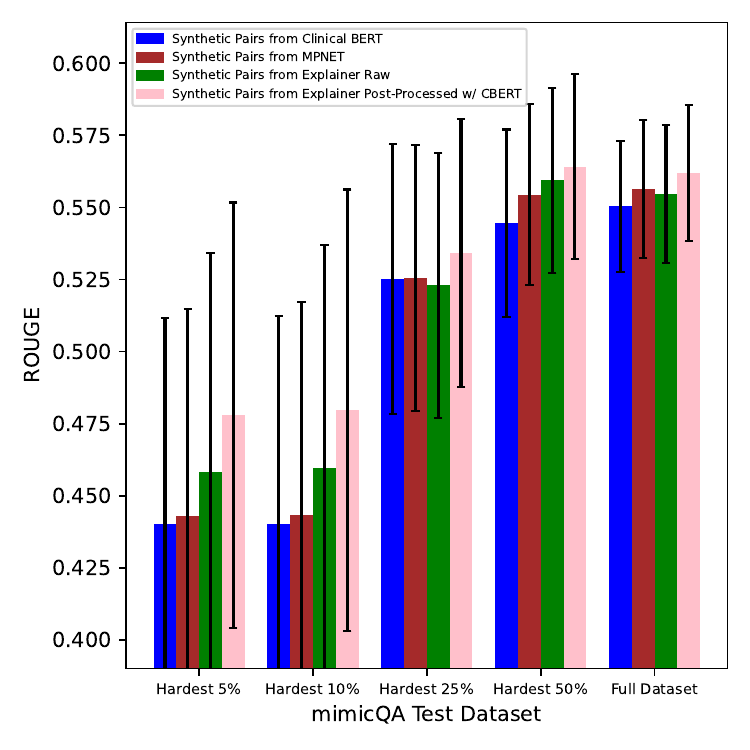}
  \vspace*{-5mm}
  \caption{ROUGE scores of GPT-4 for extractive QA using different synthetic QA pairs on hard subsets of mimicQA. CBERT is ClinicalBERT.}
  \label{fig:gpt4_mimicqa_ROUGE_base}
\end{figure}

We include plots of ROUGE score for the GPT-4 and Longformer experiments on the hard subsets of both test datasets for all methods in
Figure~\ref{fig:gpt4_emrqa_ROUGE_base}, Figure~\ref{fig:gpt4_mimicqa_ROUGE_base}, and Figure~\ref{fig:longformer_bar_results_ROUGE_base}.

\begin{figure}[h]
  \centering
  \includegraphics[width=3in]{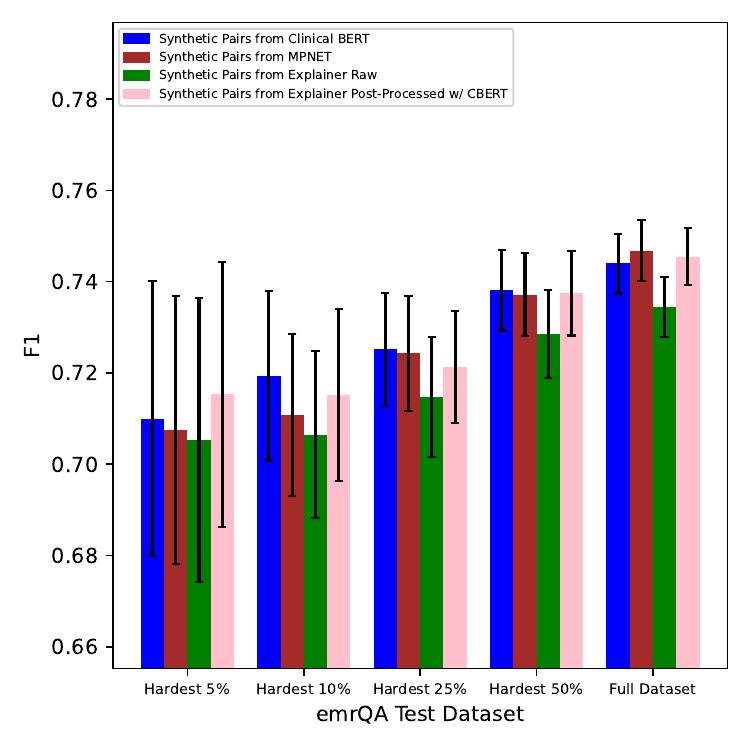}
  \vspace*{-5mm}
  \caption{F1 scores of GPT-4 for extractive QA using different synthetic QA pairs on hard subsets of emrQA relations. CBERT is ClinicalBERT.}
  \label{fig:gpt4_emrqa_f1_base}
\end{figure}

\begin{figure}[h]
  \centering
  \includegraphics[width=3in]{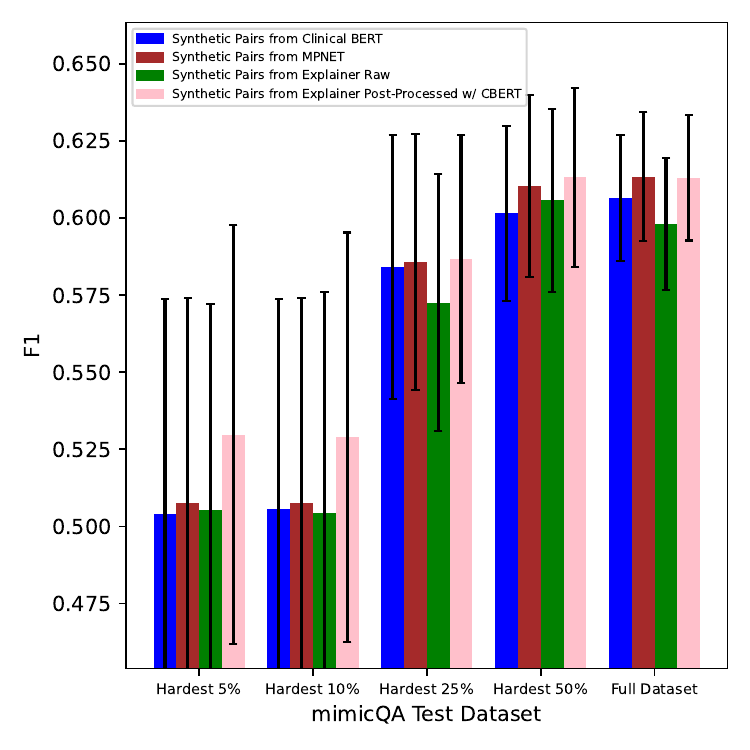}
  \vspace*{-5mm}
  \caption{F1 scores of GPT-4 for extractive QA using different synthetic QA pairs on hard subsets of mimicQA. CBERT is ClinicalBERT}
  \label{fig:gpt4_mimicqa_f1_base}
\end{figure}

\begin{figure*}[h]
  \centering
  \includegraphics[width=6.5in]{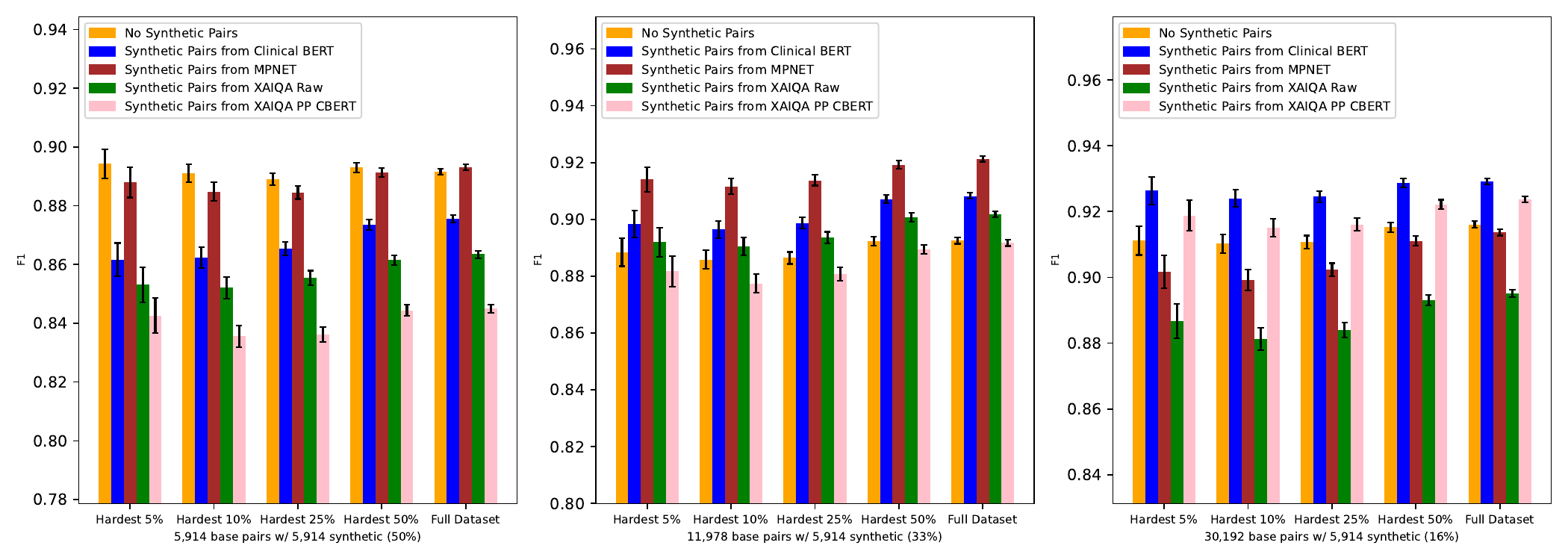}
  \caption{F1 scores of Longformer fine-tuned for extractive QA using varying ratios of base to synthetic QA pairs on emrQA relations. CBERT is ClinicalBERT.}
  \label{fig:longformer_bar_results_F1_base}
\end{figure*}

F1 score for QA balances n-gram recall and precision, rewarding models with complete predicted answers that 
are no longer than necessary.  We report F1s for the GPT-4 and Longformer experiments in 
Figure~\ref{fig:gpt4_emrqa_f1_base}, Figure~\ref{fig:gpt4_mimicqa_f1_base}, and Figure~\ref{fig:longformer_bar_results_F1_base}.

\begin{figure*}[h]
  \centering
  \includegraphics[width=6.5in]{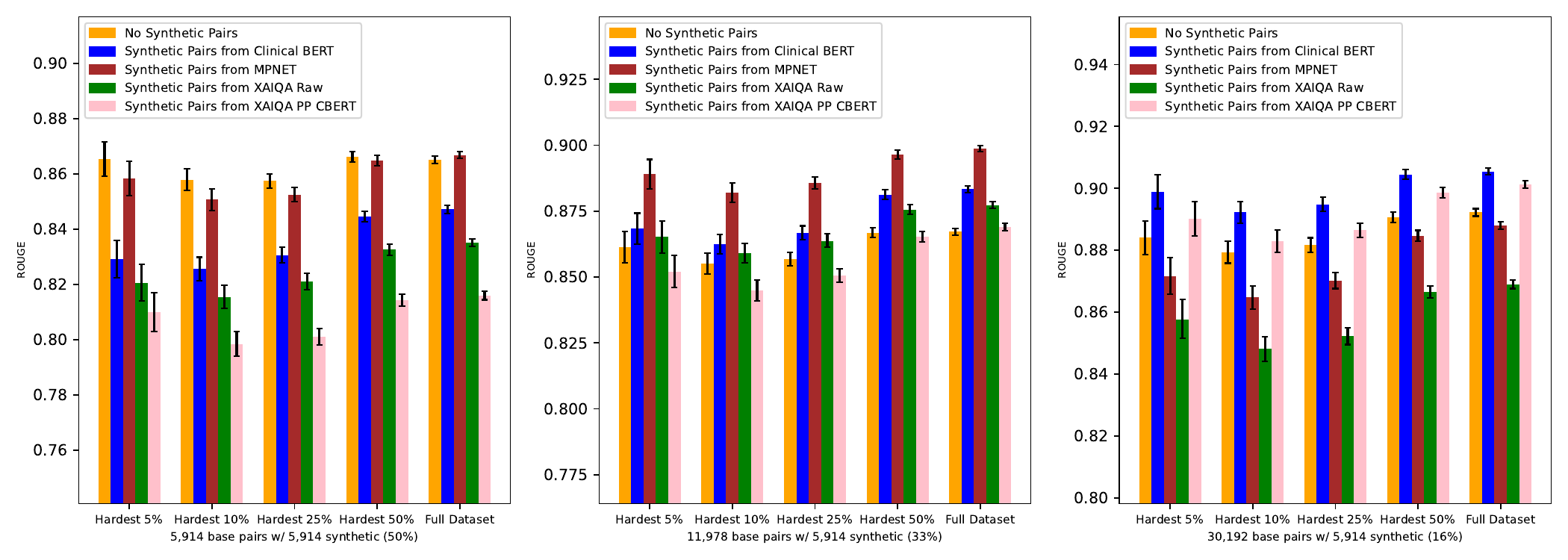}
  \caption{ROUGE scores of Longformer fine-tuned using varying ratios of base to synthetic pairs on emrQA relations.  CBERT is ClinicalBERT.}
  \label{fig:longformer_emrqa}
\end{figure*}

\end{document}